\documentclass{article}

\usepackage{PRIMEarxiv}

\usepackage[utf8]{inputenc} 
\usepackage[T1]{fontenc}    
\usepackage{hyperref}       
\usepackage{url}            
\usepackage{booktabs}       
\usepackage{amsfonts}       
\usepackage{nicefrac}       
\usepackage{microtype}      
\usepackage{lipsum}
\usepackage{fancyhdr}       
\usepackage{graphicx}       
\graphicspath{{media/}}     
\usepackage{orcidlink}
\usepackage[utf8]{inputenc}
\usepackage{amsmath,amssymb,amsfonts}
\usepackage{algorithm}
\usepackage{algorithmic}
\usepackage{booktabs}
\usepackage{graphicx}
\usepackage{amsthm}
\usepackage{xcolor}
\usepackage{hyperref}
\usepackage{subcaption}
\usepackage{enumitem}
\usepackage{bm}
\usepackage{booktabs}
\usepackage{multirow}
\usepackage{adjustbox}
\usepackage{bbm}
\hypersetup{
  colorlinks=true,
  linkcolor=blue!60!black,
  citecolor=green!50!black,
  urlcolor=blue!70!black
}
\usepackage[table]{xcolor}
\definecolor{unfairbg}{gray}{0.92}
\usepackage{wrapfig}
\newcommand{\R}{\mathbb{R}}
\newcommand{\E}{\mathbb{E}}
\newcommand{\bW}{\bm{W}}
\newcommand{\bU}{\bm{U}}
\newcommand{\bV}{\bm{V}}

\newcommand{\dW}{\Delta\bW}

\newtheorem{definition}{Definition}
\newtheorem{remark}{Remark}

\title{Filtering Memorization from Parameter-Space in Diffusion Models

}

\author{
  Yu Zhe$\dagger$ \\
  RIKEN AIP \\
  \texttt{zhe.yu@riken.jp} \\
   \And
    Yang Jiayan$\dagger$  \\
    Institute of Science Tokyo \\
  \\
     \AND
    Wei Junhao \thanks{corresponding author}  \\
    RIKEN AIP, Institute of Science Tokyo \\
    \texttt{wei.j.1721@m.isct.ac.jp} \\
   \\
    \And
    Yu-Lin Tsai \\
    University of California, Berkeley \\
\\
    \And
     Wang Chen \\
    Zhejiang University \\
 \\
}

\begin{document}
\maketitle
\renewcommand{\thefootnote}{\fnsymbol{footnote}} 
\footnotetext[1]{$\dagger$ Equal contribution.}

\begin{abstract}
Low-Rank Adaptation (LoRA) has become a widely used mechanism for customizing diffusion models, enabling users to inject new visual
concepts or styles through lightweight parameter updates. However, LoRAs can memorize training images, causing generated outputs to reproduce copyrighted or sensitive content.
This risk is particularly concerning in LoRA-sharing ecosystems, where users distribute trained LoRAs without releasing the underlying
training data. Existing approaches for mitigating memorization rely on access to the training pipeline, training data, or control over the inference process, making them difficult to apply when only the released LoRA weights are available. We propose \textbf{Base-Anchored Filtering (BAF)}, a training-free and data-free framework for post-hoc memorization
mitigation in diffusion LoRAs. BAF decomposes LoRA updates into spectral channels and measures their alignment with the principal subspace of the pretrained backbone.
Channels strongly aligned with this subspace are retained as generalizable adaptations, while weakly aligned channels are suppressed as potential carriers of memorized content. Experiments on multiple datasets and diffusion backbones demonstrate that BAF consistently reduces memorization while preserving or even improving generation quality. Our code is available in the supplementary material.

\end{abstract}

\section{Introduction}
\label{sec:intro}

Text-to-image diffusion models~\cite{ho2020denoising,rombach2022high} have become the dominant paradigm for high-fidelity image generation. Building on pretrained backbones such as Stable Diffusion~\cite{rombach2022high}, users increasingly customize generation toward specific styles, characters, or visual concepts. Low-Rank Adaptation (LoRA)~\cite{hu2022lora} enables such customization by fine-tuning a small set of low-rank weight updates from only a few reference images. This simplicity has given rise to a large-scale open ecosystem. Platforms such as Hugging Face and CivitAI host hundreds of thousands of user-uploaded LoRAs\footnote{CivitAI: \url{https://civitai.com}; Hugging Face: \url{https://huggingface.co/}}, many of which are publicly available, with direct plug-and-play usage on popular diffusion backbones.

\begin{figure}[t]
    \centering
    \includegraphics[width=\linewidth]{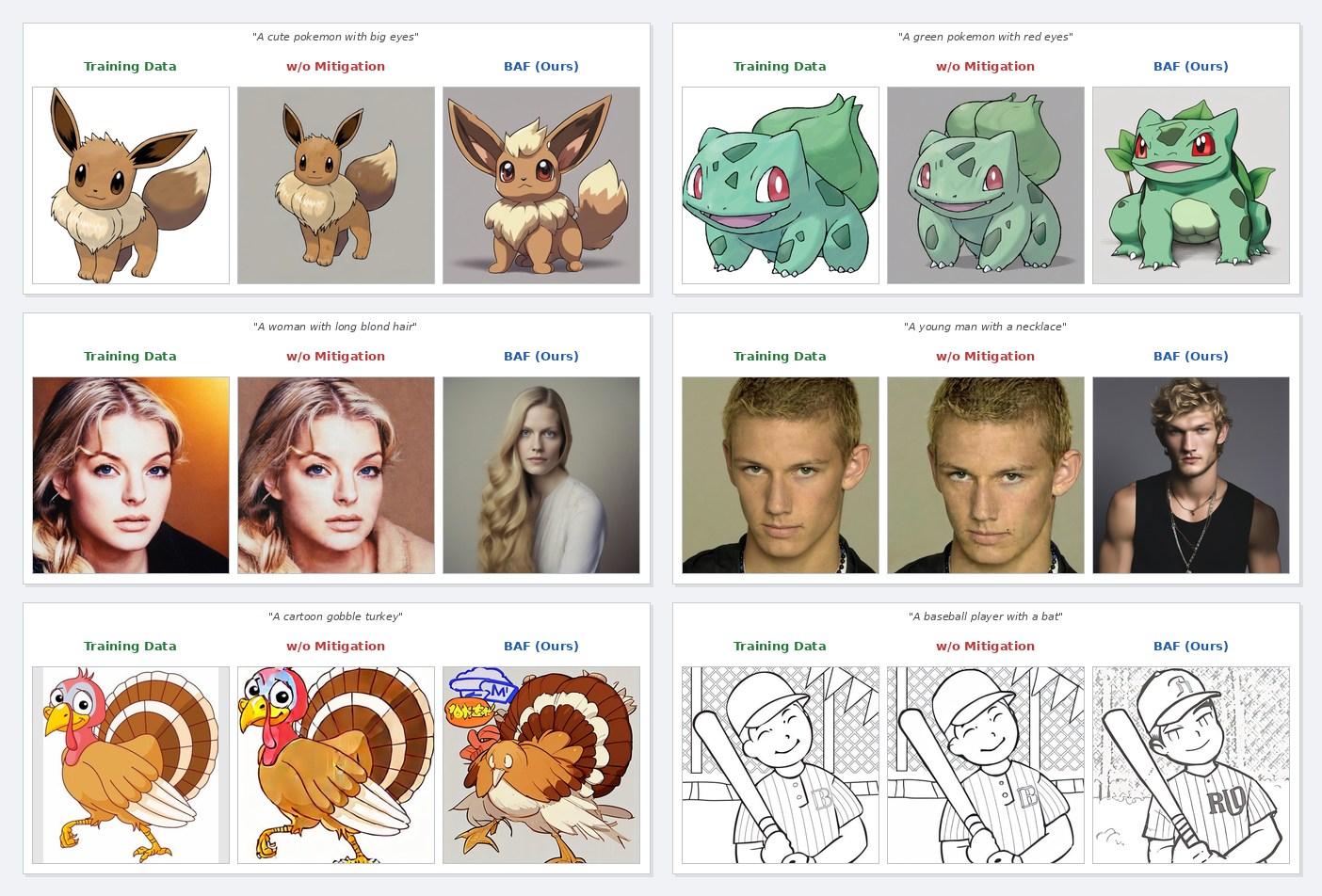}
    \caption{LoRA weight carries memorization of training data, manifested as pixel-level and object-level reproduction of training samples. BAF successfully modifies diffusion models to produce memorization-free outputs.
    }
    \label{fig:demo}
\end{figure}

While this open ecosystem enables creative flexibility, it also raises concerns regarding memorization and copyright. Particularly, memorization refers to the phenomenon whereby a model reproduces training examples or near-identical variations under certain prompts~\cite{carlini2023extracting,somepalli2023diffusion,wen2024detecting}. When the memorized content corresponds to copyrighted artwork, protected characters, or identifiable individuals, generation may constitute unauthorized reproduction. The risk is particularly acute in LoRA-sharing scenarios: downstream users typically download and deploy LoRAs without access to the fine-tuning data or optimization procedure, yet memorized content embedded in the weights can surface during normal generation (as shown in Fig.~\ref{fig:demo}). Consequently, mitigating such risks becomes important not only for platforms that host and distribute LoRA models, but also for users who wish to avoid unintended copyright or privacy violations when deploying these models.

On the other hand, existing memorization mitigation strategies are largely inapplicable in this setting. Training-time interventions~\cite{somepalli2023diffusion,tsai2025differentially} aim to prevent memorization during the training process. Yet hosting platforms and LoRA users merely receive uploaded LoRAs and do not control the fine-tuning process. Alternatively, output-based auditing methods detect memorization by comparing generated samples against known training images~\cite{carlini2023extracting,wen2024detecting}, while concept-erasure approaches remove specified learned concepts directly from model weights~\cite{gandikota2023erasing,gandikota2024unified}. Both require prior knowledge of the memorized content, which is unavailable when only the LoRA weights are uploaded. Inference-time defenses modify the sampling process to suppress memorized outputs \cite{chen2024towards}. Such approaches can be applied by users who control the inference pipeline. However, they cannot be reliably enforced at the platform level, since hosting platforms typically only distribute LoRA weights and cannot regulate downstream generation. Consequently, these approaches rely on information or control that is unavailable in LoRA-sharing ecosystems, where only the released LoRA parameters are observable. This motivates a different formulation: \emph{how can memorization be identified and filtered when only the LoRA weights are available?}


To address this question, we leverage the geometric structure of the pretrained backbone as a structural anchor. Large-scale pretraining induces a highly organized spectral structure in the network parameters~\cite{kwon2023diffusion,si2024freeu}, and meaningful task-specific variations tend to reside within low-dimensional subspaces aligned with the principal directions of the pretrained weights~\cite{aghajanyan2021intrinsic}. This discovery has led to a key insight.
\begin{center}
    \emph{Generalizable adaptation follows dominant spectral directions, whereas memorization introduces weakly aligned instance-specific directions.}
\end{center}
Empirically, we decompose LoRA updates via singular value decomposition (SVD) and measure the alignment of spectral channels with the pretrained model's principal subspace. We then observe that generation from high-alignment channels preserves the intended adaptation while exhibiting minimal memorization, whereas generation from low-alignment channels reproduces memorized training images at significantly elevated rates. This separation suggests that memorization is predominantly concentrated in spectral directions that are weakly aligned with the pretrained backbone, leaving a detectable structural signature in parameter space.

Building on this observation, we introduce \textbf{BAF} (Base-Anchored Filtering), a training-free and data-free framework for post-hoc memorization mitigation in diffusion LoRAs. BAF decomposes each LoRA update into spectral channels via SVD and measures their alignment with the principal subspace of the pretrained backbone. Channels with strong alignment are retained as generalizable adaptations, whereas weakly aligned channels are down-weighted or removed as potential memorization carriers. By operating directly in parameter space, BAF mitigates memorization without requiring access to fine-tuning data, knowledge of memorized content, or additional optimization, making it applicable to both hosting platforms and downstream LoRA users.

Our contributions are as follows:
\begin{enumerate}
    \item We formulate the problem of post-hoc memorization mitigation for diffusion LoRAs under strictly data-inaccessible constraints.
    
    \item We propose and validate a geometric insight distinguishing generalizable adaptation from memorization via spectral alignment with the pretrained principal subspace.
    
    \item We introduce Base-Anchored Filtering (BAF), a training-free and data-free framework that selectively suppresses memorization-related spectral directions in parameter space.
    
    \item We demonstrate that BAF effectively reduces memorization while preserving benign generative capability, achieving better trade-offs compared to existing memorization suppression strategies.

\end{enumerate}

\section{Related work}
\label{sec: related work}

\subsection{Memorization in diffusion model}
Memorization in diffusion models refers to the tendency of a model to reproduce training images or near-identical variants during generation. Early studies~\cite{somepalli2023diffusion,carlini2023extracting} showed that text-to-image systems such as Stable Diffusion replicate a non-trivial portion of their training data. Subsequent work investigated the mechanisms behind memorization. \cite{gu2023memorization} analyzed the effects of dataset size, duplication rate, and model capacity, and observed that in LoRA fine-tuning, memorization strength increases with the rank of the adaptation matrices. At a different granularity, \cite{wen2024detecting} linked memorization to abnormal text-conditional prediction magnitudes and proposed token-level attribution to identify triggering words. \cite{ren2024unveiling} localized memorization effects to cross-attention patterns, while \cite{chen2024exploring} distinguished between \emph{global} and \emph{local} memorization, showing that the latter is harder to detect and mitigate. From a geometric perspective, \cite{ross2024geometric} connected memorization to regions of near-zero Local Intrinsic Dimensionality (LID).

\subsection{Memorization Detection and Mitigation}
\label{sec:related_mem}
Existing work on mitigating memorization can be broadly categorized into training-time, inference-time, and parameter-space approaches.

\emph{Training-time approaches} reduce memorization by modifying the training data or optimization process. Data deduplication~\cite{openai2022dalle2,kandpal2022deduplicating} removes near-duplicate images from training sets, while differential privacy methods~\cite{tsai2025differentially,dockhorn2022differentially} limit the influence of individual samples through gradient clipping and noise injection. However, these approaches require access to the training process. In LoRA-sharing ecosystems, hosting platforms typically receive only the final LoRA weights and cannot influence how the adapter was trained.

\emph{Inference-time approaches} intervene during sampling. Random Token Addition (RTA)~\cite{somepalli2023understanding} perturbs prompts to disrupt memorization-triggering conditioning signals. \cite{wen2024detecting} optimizes prompt embeddings to reduce text-conditional prediction magnitudes, steering generation away from memorized samples. \cite{jain2025classifier} studies how classifier-free guidance amplifies memorization and proposes modifying guidance within the attraction basin of the denoising trajectory. \cite{chen2024towards} introduces Anti-Memorization Guidance (AMG), which detects replication during sampling and applies corrective guidance, partially relying on nearest-neighbor search over training images. However, these methods leave memorized content intact in model parameters and only suppress its retrieval during sampling.

\emph{Parameter-space approaches} attempt to mitigate memorization by directly editing model weights. Concept-erasure methods~\cite{gandikota2023erasing,gandikota2024unified} fine-tune models to remove specified concepts, requiring the targets to be known. \cite{hintersdorf2024finding} identifies neurons in cross-attention value layers through outlier activations using memorized prompts. Similarly, \cite{chavhan2024memorized} applies Wanda pruning to remove weights activated by memorized prompts in feed-forward layers. Despite operating in parameter space, these methods still rely on prior knowledge of memorized prompts or concepts. In the LoRA-sharing scenario we consider, the hosting platform or LoRA users only have access to the released LoRA weights and cannot determine which training images or sensitive content may have been memorized during fine-tuning.

\section{Preliminaries}
\label{sec:prelim}
\paragraph{Low-Rank Adaptation.}
Low-Rank Adaptation (LoRA)~\cite{hu2022lora} fine-tunes a pretrained model by 
adding trainable low-rank updates to selected layers.
For a pretrained weight matrix $\bW_* \in \R^{m \times n}$, LoRA introduces 
$B \in \R^{m \times r}$ and $A \in \R^{r \times n}$ with rank 
$r \ll \min(m, n)$, and the adapted weight is $\bW = \bW_* + \dW$ where 
$\dW \triangleq BA$ is the additive weight update.
To analyze the internal structure of LoRA updates, we turn to their spectral 
decomposition.

\paragraph{Spectral decomposition of LoRA updates.}
The singular value decomposition (SVD) of a rank-$r$ LoRA update is $\dW = \sum_{i=1}^{r} \sigma_i \, \bm{u}_i \bm{v}_i^\top,$
where $\sigma_1 \geq \cdots \geq \sigma_r > 0$ are singular values and
$\bm{u}_i \in \R^m$, $\bm{v}_i \in \R^n$ are left/right singular vectors.
Each rank-1 term $\sigma_i \bm{u}_i \bm{v}_i^\top$ is a \emph{spectral channel}: 
it maps an input direction $\bm{v}_i$ to an output direction $\bm{u}_i$ with 
magnitude $\sigma_i$.

\paragraph{Principal subspace of the pretrained model.}
Let $\bW_* \in \mathbb{R}^{m\times n}$ denote the pretrained weight at the same layer.
We compute its singular value decomposition (SVD):
\[
\bW_* = \bU_* \bm{\Sigma}_* \bV_*^\top
= \sum_{j=1}^{\min(m,n)} s_j \, \bm{u}_j^* {\bm{v}_j^*}^\top ,
\]
where $\bU_*=[\bm{u}_1^*,\ldots,\bm{u}_m^*]$ and 
$\bV_*=[\bm{v}_1^*,\ldots,\bm{v}_n^*]$ are orthonormal matrices and 
$\bm{\Sigma}_*=\mathrm{diag}(s_1,\ldots,s_{\min(m,n)})$ contains the singular values.

Let $\bU_{*,K}$ and $\bV_{*,K}$ denote the matrices containing the first $K$
left and right singular vectors of $\bW_*$. 
We define the orthogonal projection operators onto the corresponding
principal subspaces as
\[
\bm{P}_L = \bU_{*,K}\bU_{*,K}^\top,
\qquad
\bm{P}_R = \bV_{*,K}\bV_{*,K}^\top.
\]

\paragraph{Problem Setup.}

We consider a setting where only a released LoRA adapter is available. Each LoRA weight $\dW$ is designed to be applied to a publicly available pretrained base model $\bW_*$. The base model weights $\bW_*$ are known, and the LoRA update $\dW$ is observable. However, the fine-tuning data, training procedure, and hyperparameters used to produce the LoRA are unknown, and there is no prior knowledge of what content may have been memorized. The goal is to produce a filtered update $\dW_{\mathrm{filtered}}$ that (i) preserves generalizable adaptation encoded in the LoRA
(e.g., artistic style or character design) while (ii) suppresses memorized training content (e.g., near-verbatim copies of copyrighted images).

We note that this setting reflects current diffusion ecosystems in which LoRA adapters are widely shared and downloaded. Platforms such as Hugging Face and CivitAI host large collections of
user-uploaded LoRAs, while downstream users may deploy these adapters without access to the original training data.
In such scenarios, only the released LoRA parameters are observable, and filtering must rely solely on $\dW$ to separate generalized adaptation from memorization.

\section{Main Method}
\label{sec:method}

We propose \textbf{BAF} (Base-Anchored Filtering), a data-free framework for mitigating memorization in LoRA weights using only parameter-space structure.
Our key insight is that spectral channels aligned with the pretrained model's principal subspace tend to encode \emph{generalizable adaptation}, while weakly aligned channels are more likely to capture \emph{memorized content}.

Section~\ref{sec:method} proceeds as follows. We first define an \emph{anchoring score} to measure spectral alignment and empirically validate the hypothesis via a directional ablation experiment (Section~\ref{sec:empirical}). We then present the complete BAF filtering algorithm (Section~\ref{sec:algorithm}). Finally, we introduce an \emph{adaptive strategy} for selecting the principal-subspace dimension $K$ across different architectures, based on an energy-adaptive rule (Section~\ref{sec:kchoice}).

\subsection{Empirical Motivation}
\label{sec:empirical}

To validate our insight directly, we first define a quantitative measure of spectral alignment.

\begin{definition}[Anchoring score]
\label{def:anchoring}
For the $i$-th spectral channel with left and right singular vectors 
$\bm{u}_i \in \R^m$ and $\bm{v}_i \in \R^n$, the \emph{anchoring score} with 
respect to the pretrained model $\bW_*$'s top-$K$ principal subspace is defined as
\begin{equation}
  a_i = \|\bm{P}_L \bm{u}_i\|^2 \cdot \|\bm{P}_R \bm{v}_i\|^2 \in [0,1],
  \label{eq:anchoring}
\end{equation}
where $\bm{P}_L$ and $\bm{P}_R$ are the orthogonal projections onto the 
pretrained left and right principal subspaces.
\end{definition}

\begin{remark}
The anchoring score admits a simple geometric interpretation. Since
singular vectors are unit vectors, the projection norm satisfies
$\|\bm{P}_L \bm{u}_i\|^2 = \cos^2\theta_i^L$, where $\theta_i^L$ is the
angle between $\bm{u}_i$ and the pretrained left subspace.
Thus
\[a_i = \cos^2\theta_i^L \cdot \cos^2\theta_i^R \in [0,1]\]
measures the bilateral alignment of the $i$-th spectral channel with the
pretrained subspaces. Channels with large $a_i$ lie largely within the
pretrained directions, whereas channels with small $a_i$ introduce
directions that are largely orthogonal to the pretrained model.

This geometric view is conceptually consistent with recent analyses of
geometric memorization~\cite{ross2024geometric}, which suggest that
memorized samples tend to lie in regions of near-zero Local Intrinsic
Dimensionality (LID) in data space, corresponding to isolated pockets of
the data manifold. Such instance-specific directions are weakly
represented by the dominant structure of the data distribution and are
therefore less likely to align with the pretrained principal subspace,
resulting in low anchoring scores. From this perspective, the anchoring
score can be interpreted as a parameter-space analogue of LID, providing
a complementary view of the memorization from the spectral structure of
model parameters.

\end{remark}

\begin{definition}[Random-projection baseline]
\label{def:null}
If $\bm{u}$ and $\bm{v}$ are independently drawn uniformly from the unit 
spheres in $\R^m$ and $\R^n$, then the expected anchoring score is
\begin{equation}
  a_{\mathrm{null}} \triangleq \E[a(\bm{u}, \bm{v})]
  = \frac{K}{m} \cdot \frac{K}{n}
  = \frac{K^2}{mn}.
  \label{eq:null}
\end{equation}
\end{definition}

\begin{remark}
The baseline $a_{\mathrm{null}}$ represents the expected anchoring score of a channel that carries no structured relationship to the pretrained
model. Intuitively, if a spectral channel does not reuse the pretrained directions, its singular vectors behave like arbitrary directions in
weight space. Modeling them as random unit vectors provides a natural reference point for alignment. Channels with $a_i \gg a_{\mathrm{null}}$ are strongly anchored to the pretrained subspace, whereas channels with scores close to $a_{\mathrm{null}}$ exhibit no more alignment than random directions.
\end{remark}
\paragraph{Directional ablation experiment.}
We validate our insight via a directional ablation that isolates high- and low-anchoring components of a memorizing LoRA. To create a high-memorization setting, we augment the training data with 50 canary images duplicated 50 times, producing a \emph{High-mem LoRA}, where memorization signals are amplified. As a control, we also train LoRAs without duplicated canary images, referred to as \emph{Low-mem LoRA}, representing standard training without explicit memorization amplification and serving as a low-memorization baseline.

For each layer, we compute the SVD of $\Delta W$ and split channels based on the layer-wise baseline $a_{\mathrm{null}}$. Since the analysis is applied independently to each layer, we omit the layer index for simplicity. We then construct two LoRA variants from \emph{High-mem LoRA}: a \emph{high-align-only} LoRA that keeps channels with a higher anchoring score than the random-projection baseline (i.e, $a_i \ge a_{\mathrm{null}}$) and zeros the rest, and a \emph{low-align-only} LoRA that keeps channels with $a_i < a_{\mathrm{null}}$ and removes the high-alignment components. 

\paragraph{Metrics.}
\emph{Memorization} is measured by cosine similarity in CLIP embedding space between generated and canary training images. \emph{Generation quality} is measured by aesthetic score~\cite{laion_aesthetics}.

\paragraph{Results.}
\begin{figure}[t]
  \centering
  \includegraphics[width=\linewidth]{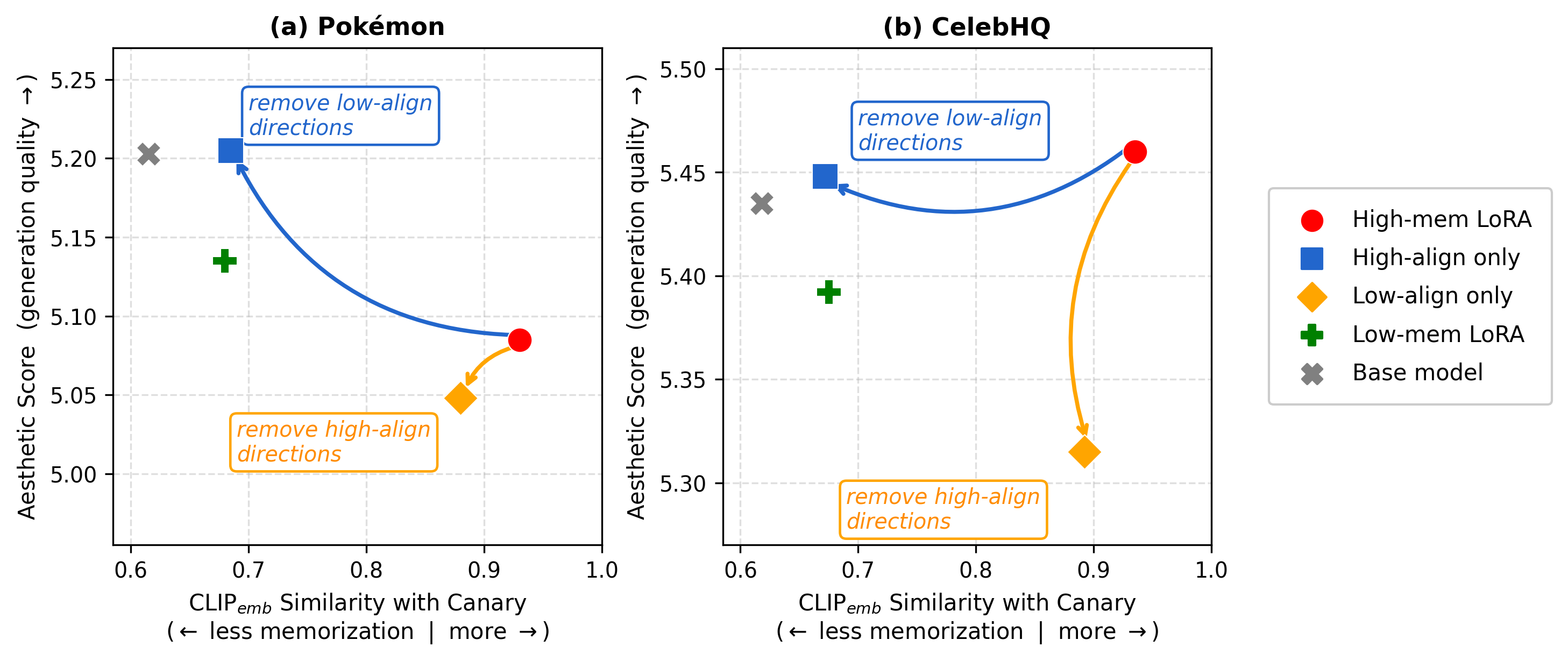}
  \caption{%
    \textbf{Directional ablation supports the anchoring hypothesis.}
    From a memorizing LoRA (\textcolor{red}{$\bullet$}), removing low-alignment 
    channels yields near-base memorization with the best aesthetic quality 
    (\textcolor{blue}{$\blacksquare$}).
    Removing high-alignment channels preserves memorization but lowers quality 
    (\textcolor{orange}{$\blacklozenge$}).
    The clean LoRA (\textcolor{green!60!black}{$+$}) and base model ($\times$) 
    are references.
    Results are consistent across Pokémon (a) and CelebHQ (b).
  }
  \label{fig:verification}
\end{figure}

Figure~\ref{fig:verification} shows consistent trends on both datasets. For the High-mem LoRA, retaining only high-alignment channels reduces memorization to the base-model level while preserving aesthetic quality.  In contrast, retaining only low-alignment channels maintains high memorization and degrades generation quality. These results suggest that LoRAs with highly-aligned channels mainly encode generalizable adaptation, whereas lowly-aligned channels tend to carry memorized content with limited contribution to visual quality. Detailed experimental settings are provided in the appendix A.

\subsection{The BAF Algorithm}
\label{sec:algorithm}

Motivated by the above experiment, BAF applies anchoring-based gating to each layer of a 
single LoRA update. We define two gating modes.

\paragraph{BAF-Hard.}
Binary gating with threshold $a_{\mathrm{null}}$:
\begin{equation}
g_{\mathrm{hard}}(a_i) = \mathbbm{1}\!\left[a_i \ge a_{\mathrm{null}}\right].
\label{eq:hard_gate}
\end{equation}
BAF-Hard explicitly removes channels below the anchoring threshold.

\paragraph{BAF-Soft.}
While BAF-Hard may drastically reduce memorization, it could inadvertently remove useful channels near the anchoring threshold. BAF-Soft instead suppresses rather than removes them via continuous gating with 
exponent $\alpha > 0$:
\begin{equation}
  g_{\mathrm{soft}}(a_i) = a_i^\alpha.
  \label{eq:soft_gate}
\end{equation}
Larger $\alpha$ yields stronger suppression; $\alpha = 0$ recovers the 
unfiltered LoRA update.

\paragraph{Filtered update.}
BAF produces the filtered LoRA update at each layer:
\begin{equation}
  \dW_{\mathrm{BAF}} =
  \sum_{i=1}^{r}
  \sigma_i \cdot g\!\left(a_i\right)\cdot
  \bm{u}_i \bm{v}_i^\top,
  \label{eq:baf_filter}
\end{equation}
where the gating function $g(\cdot)$ depends on the method of choice with the final adapted model being $\bW = \bW_* + 
\dW_{\mathrm{BAF}}$.
The full algorithm is given in the supplementary material (Appendix B).

\subsection{Energy-Adaptive Principal Subspace Selection}
\label{sec:kchoice}

Selecting the principal-subspace dimension $K$ is important for applying BAF across different architectures. Pretrained layers often exhibit heterogeneous spectral decay: some concentrate most energy in a few dominant singular directions, while others distribute energy across many directions. As a result, the effective principal subspace dimension varies across layers, making a globally fixed $K$ suboptimal.

\paragraph{Energy-adaptive selection rule.}
We therefore select $K$ based on the captured spectral energy. For each layer \footnote{since the following adaptive strategy is applied independently to each layer, we omit the layer index}:
\begin{equation}
  K =
  \min\left\{k \in \{1,\dots,\min(m,n)\} :
  \frac{\sum_{j=1}^{k} s_j^2}{\sum_{j=1}^{\min(m,n)} s_j^2}
  \ge \tau_{\mathrm{energy}} \right\},
  \label{eq:adaptive_K}
\end{equation}
where $s_j$ are the singular values of $\bW_*$ and 
$\tau_{\mathrm{energy}} \in (0,1)$ is an energy threshold.
This rule adapts $K$ to the intrinsic spectral structure of each layer:
layers with sharp spectral decay yield smaller subspaces, while flatter spectra
retain larger ones.

\section{Experiments}
\label{sec:experiments}
\paragraph{Setup.} We evaluate on two diffusion backbones spanning different scales, Stable Diffusion v1.5~\cite{rombach2022high} and SDXL~\cite{podell2023sdxl}, using LoRAs trained on the U-Net attention projection layers. 
In the main paper, experiments are conducted on three datasets: \emph{Pokémon}~\cite{pinkney2022Pokemon}, a stylized character dataset of about 800 images with BLIP-generated captions, and \emph{CelebA-HQ}~\cite{karras2018progressive}, where we sample subsets to simulate portrait-domain LoRAs, and \emph{LAION Art}~\cite{somepalli2023understanding}, constructed by retrieving images with high CLIP similarity to art-related prompts and filtering with an aesthetic predictor (about 1{,}000 images). Experimental results on Stable Diffusion v2.1, and detailed training settings are reported in the supplementary material (Appendix C) due to space constraints.

\paragraph{Scenario.} Since memorization events are difficult to control under standard training conditions, prior work studies memorization under controlled settings by increasing the frequency of selected training samples during training~\cite{wen2024detecting}. Following this intuition, we simulate memorization during LoRA training by duplicating a set of canary images $\{(x_k, y_k)\}_{k=1}^{N_c}$ in the training
data, where each canary image is repeated $D$ times. This controlled setup allows us to reliably induce memorization in the resulting LoRA updates, which is otherwise relatively weak under standard training conditions~\cite{gu2023memorization}. In our main experiments, we use a fixed duplication factor ($D=50, N_c=50$) to create strongly memorizing LoRAs, while Sec.~\ref{sec:ablations} further studies different memorization strengths. This setup also reflects a realistic risk in model-sharing ecosystems: an uploader could intentionally amplify memorization of copyrighted images during training and release the resulting LoRA, causing deployed models to reproduce protected content with high fidelity.

\paragraph{Evaluation metrics.}
We use SSCD~\cite{pizzi2022self} for measuring memorization, a cosine similarity metric in a copy-detection embedding space. We report \textbf{targeted SSCD}, defined as the similarity between each generated image and its corresponding training image (duplicated canary). Additional memorization metrics (normalized $\ell_2$ distance and CLIP embedding distance) are provided in the Appendix D. For generation quality and prompt fidelity, we evaluate text-image alignment using CLIP score~\cite{radford2021learning} and visual quality using aesthetic score~\cite{schuhmann2022laion}. Additional metrics (LPIPS~\cite{zhang2018unreasonable}, ImageReward~\cite{xu2023imagereward}) are reported in the appendix D.

\paragraph{Baselines.}
We compare against representative methods spanning two mitigation categories. \emph{Training-time.}
\textbf{Deduplication}~\cite{somepalli2023diffusion,kandpal2022deduplicating} removes near-duplicate images before fine-tuning, and \textbf{DP-LoRA}~\cite{tsai2025differentially}
applies differentially private optimization during LoRA training. These methods require access to the training pipeline, which is unavailable in our setting where only released LoRA weights are observable, but we include them as reference baselines.

\emph{Inference-time.}
\textbf{RTA}~\cite{somepalli2023understanding}, \textbf{Prompt Optim.}~\cite{wen2024detecting},
\textbf{CFG-Reverse}~\cite{jain2025classifier}, and \textbf{AMG}~\cite{chen2024towards} modify the sampling process to reduce memorization during generation.
These methods (besides AMG) operate without requiring access to the training pipeline, making them comparable to our approach in terms of available information and capabilities. We therefore include them as baselines in our evaluation. Among these methods, AMG additionally relies on access to training images. Although this assumption does not hold in our setting, we report AMG as a strong data-dependent baseline to demonstrate the difference between data-free and data-dependent mitigation approaches. For BAF, we report BAF-Hard and BAF-Soft ($\alpha=1$) with $\tau_\mathrm{energy}=0.85$ by default.

\subsection{Main experiments}

\begin{table}[t]
\centering
\scriptsize
\setlength{\tabcolsep}{2.8pt}
\renewcommand{\arraystretch}{1.05}

\caption{\textbf{Evaluation of memorization mitigation and generation quality when LoRA trained on Pokémon or CelebA-HQ.}
\colorbox{unfairbg}{Shaded rows} refer to methods that are incomparable to our proposal.
Best and second-best among comparable (non-shaded) methods are \textbf{bolded} and \underline{underlined}, respectively.}

\label{tab:scenarioA_Pokémon_celeb_clip}

\scalebox{0.83}{%
\begin{tabular}{l ccc ccc ccc ccc}
\toprule
\multirow{3}{*}{Method}
& \multicolumn{6}{c}{\textbf{Pokémon}}
& \multicolumn{6}{c}{\textbf{CelebA-HQ}} \\
\cmidrule(lr){2-7} \cmidrule(lr){8-13}
& \multicolumn{3}{c}{SD1.5}
& \multicolumn{3}{c}{SDXL}
& \multicolumn{3}{c}{SD1.5}
& \multicolumn{3}{c}{SDXL} \\
\cmidrule(lr){2-4} \cmidrule(lr){5-7}
\cmidrule(lr){8-10} \cmidrule(lr){11-13}
& SSCD$\downarrow$ & AES$\uparrow$ & CLIP$\uparrow$
& SSCD$\downarrow$ & AES$\uparrow$ & CLIP$\uparrow$
& SSCD$\downarrow$ & AES$\uparrow$ & CLIP$\uparrow$
& SSCD$\downarrow$ & AES$\uparrow$ & CLIP$\uparrow$ \\
\midrule

W/O mitigation
& 0.76 & 5.08 & 33.20
& 0.74 & 4.87 & 33.16
& 0.93 & 5.46 & 29.54
& 0.84 & 4.72 & 29.03 \\

\rowcolor{unfairbg}
Deduplication
& 0.65 & 5.04 & 33.26
& 0.63 & 4.74 & 32.83
& 0.53 & 5.47 & 29.49
& 0.56 & 5.05 & 28.63 \\

\rowcolor{unfairbg}
DP-LoRA
& 0.60 & 4.85 & 32.00
& 0.66 & 4.40 & 30.50
& 0.62 & 5.25 & 28.90
& 0.70 & 4.60 & 28.40 \\

\midrule

RTA
& 0.72 & \underline{5.12} & 32.22
& 0.68 & 5.10 & 30.02
& 0.74 & 5.34 & 29.11
& 0.81 & 4.91 & 28.27 \\

Prompt Optim.
& 0.66 & 4.95 & 31.50
& 0.71 & 4.60 & 29.80
& 0.68 & 5.10 & 28.85
& 0.74 & 4.70 & 27.90 \\

CFG-Reverse
& 0.79 & 5.06 & \underline{33.24}
& 0.78 & 4.89 & 33.39
& 0.71 & \underline{5.53} & 29.65
& 0.75 & 5.28 & 30.02 \\

\rowcolor{unfairbg}
AMG
& 0.50 & 5.18 & 33.05
& 0.56 & 5.03 & 32.82
& 0.40 & 5.52 & 29.72
& 0.63 & 5.26 & 29.84 \\

\midrule

BAF-Soft
& \underline{0.57} & \textbf{5.43} & \textbf{33.40}
& \underline{0.62} & \underline{5.71} & \textbf{34.27}
& \underline{0.52} & \textbf{5.67} & \textbf{31.45}
& \underline{0.68} & \textbf{6.13} & \textbf{32.45} \\

BAF-Hard
& \textbf{0.54} & 5.02 & 32.38
& \textbf{0.60} & \textbf{5.49} & \underline{34.23}
& \textbf{0.44} & 5.51 & \underline{31.24}
& \textbf{0.67} & \underline{6.10} & \underline{32.36} \\

\bottomrule
\end{tabular}%
}
\end{table}

\begin{table}[t]
\centering
\scriptsize
\setlength{\tabcolsep}{3.0pt}
\renewcommand{\arraystretch}{1.05}
\caption{\textbf{Evaluation of memorization mitigation and generation quality when LoRA trained on LAION-Art }.
\colorbox{unfairbg}{Shaded rows} refer to methods that are incomparable to our proposal. Best and second-best among fairly comparable (non-shaded) methods are \textbf{bolded} and \underline{underlined}, respectively.}

\label{tab:scenarioA_LAION_clip}

\scalebox{0.83}{%
\begin{tabular}{l ccc ccc}
\toprule
\multirow{2}{*}{Method}
& \multicolumn{3}{c}{SD1.5}
& \multicolumn{3}{c}{SDXL} \\
\cmidrule(lr){2-4} \cmidrule(lr){5-7}

& SSCD$\downarrow$ & AES$\uparrow$ & CLIP$\uparrow$
& SSCD$\downarrow$ & AES$\uparrow$ & CLIP$\uparrow$ \\

\midrule

W/O mitigation
& 0.83 & 5.78 & 31.45
& 0.79 & 5.42 & 31.18 \\

\rowcolor{unfairbg}
Deduplication
& 0.71 & 5.70 & 31.30
& 0.73 & 5.31 & 30.95 \\

\rowcolor{unfairbg}
DP-LoRA
& 0.69 & 5.55 & 30.90
& 0.71 & 5.18 & 30.42 \\

\midrule

RTA
& 0.76 & \underline{5.81} & 31.32
& 0.73 & 5.45 & 30.98 \\

Prompt Optim.
& 0.72 & 5.63 & 31.05
& 0.74 & 5.28 & 30.70 \\

CFG-Reverse
& 0.82 & 5.75 & 31.58
& 0.80 & 5.47 & 31.44 \\

\rowcolor{unfairbg}
AMG
& 0.46 & 5.82 & 31.61
& 0.48 & 5.51 & 31.52 \\

\midrule

BAF-Soft
& \underline{0.52} & \textbf{5.97} & \textbf{33.52}
& \underline{0.54} & \textbf{6.01} & \textbf{34.23} \\

BAF-Hard
& \textbf{0.49} & 5.77 & \underline{33.47}
& \textbf{0.52} & \underline{5.89} & \underline{34.19} \\

\bottomrule
\end{tabular}%
}
\end{table}

Table~\ref{tab:scenarioA_Pokémon_celeb_clip} and Table~\ref{tab:scenarioA_LAION_clip} report results across three datasets (Pokémon, CelebA-HQ, and LAION-Art) and two diffusion backbones
(SD1.5 and SDXL). Across all settings, both BAF variants consistently achieve the lowest SSCD among \emph{fairly comparable} methods, and even achieve better results than several methods that require additional information. We note that AMG achieves lower SSCD than the proposal because it suppresses memorization through similarity detection against the training set. However, this mechanism requires access to training images and relies on nearest-neighbor retrieval in the training embedding space during generation, introducing additional data dependency and computational overhead. 

Importantly, BAF reduces memorization without sacrificing generation quality. Due to space constraints, we provide visualizations of the generated images with or without BAF in the Appendix E. Across most settings, both AES and CLIP scores improve compared to the
unfiltered LoRA. This suggests that weakly anchored channels may correspond to noisy parameter directions, and removing them can improve generation quality. BAF-Hard and BAF-Soft provide different trade-offs. BAF-Hard achieves stronger memorization suppression, while BAF-Soft preserves slightly better generation quality. In practice, BAF-Soft provides a more balanced trade-off, and we use it as the default configuration in the following analyses.

\subsection{Ablation studies}
\label{sec:ablations}
\begin{figure}[t]
  \centering
  \includegraphics[width=0.6 \linewidth]{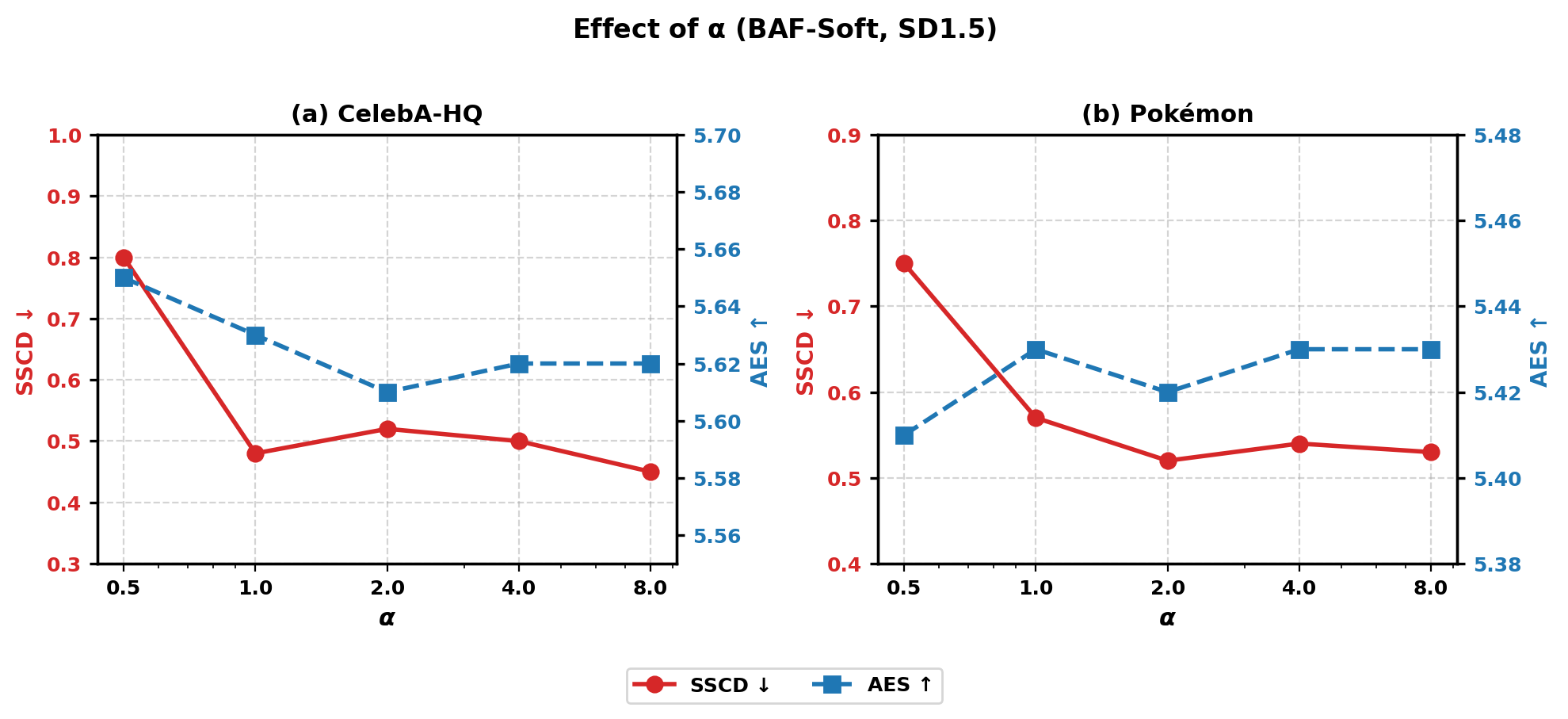}
  \caption{%
    \textbf{Effect of $\alpha$ in BAF-Soft on CelebA-HQ (a) and Pokémon (b), both trained on SD1.5.}
    We measure SSCD (red) and AES (blue).
  }
  \label{fig:alpha_sensitivity}
\end{figure}

\paragraph{Effect of $\alpha$ in BAF-Soft.}
Figure~\ref{fig:alpha_sensitivity} studies the effect of the scaling parameter $\alpha$ in BAF-Soft.
Across both CelebA-HQ and Pokémon, the most notable drop in SSCD occurs when $\alpha$ increases from $0.5$ to $1$, indicating that even a moderate scaling factor is sufficient to substantially suppress memorization.
Beyond $\alpha = 1$, SSCD generally decreases or stabilizes with only minor fluctuations, suggesting that the benefit of further increasing $\alpha$ diminishes.
Importantly, the aesthetic score (AES) remains nearly constant across the entire range of $\alpha$ on both datasets, with a variation of less than $1\%$.
These results suggest that $\alpha$ primarily controls the strength of memorization suppression, while having minimal impact on generation quality, and that $\alpha = 1$ offers a favorable trade-off between the two objectives.

\paragraph{Effect of $\tau_\text{energy}$ in BAF-Soft and visualization.}

\begin{figure}[t]
  \centering
  \includegraphics[width=0.6\linewidth]{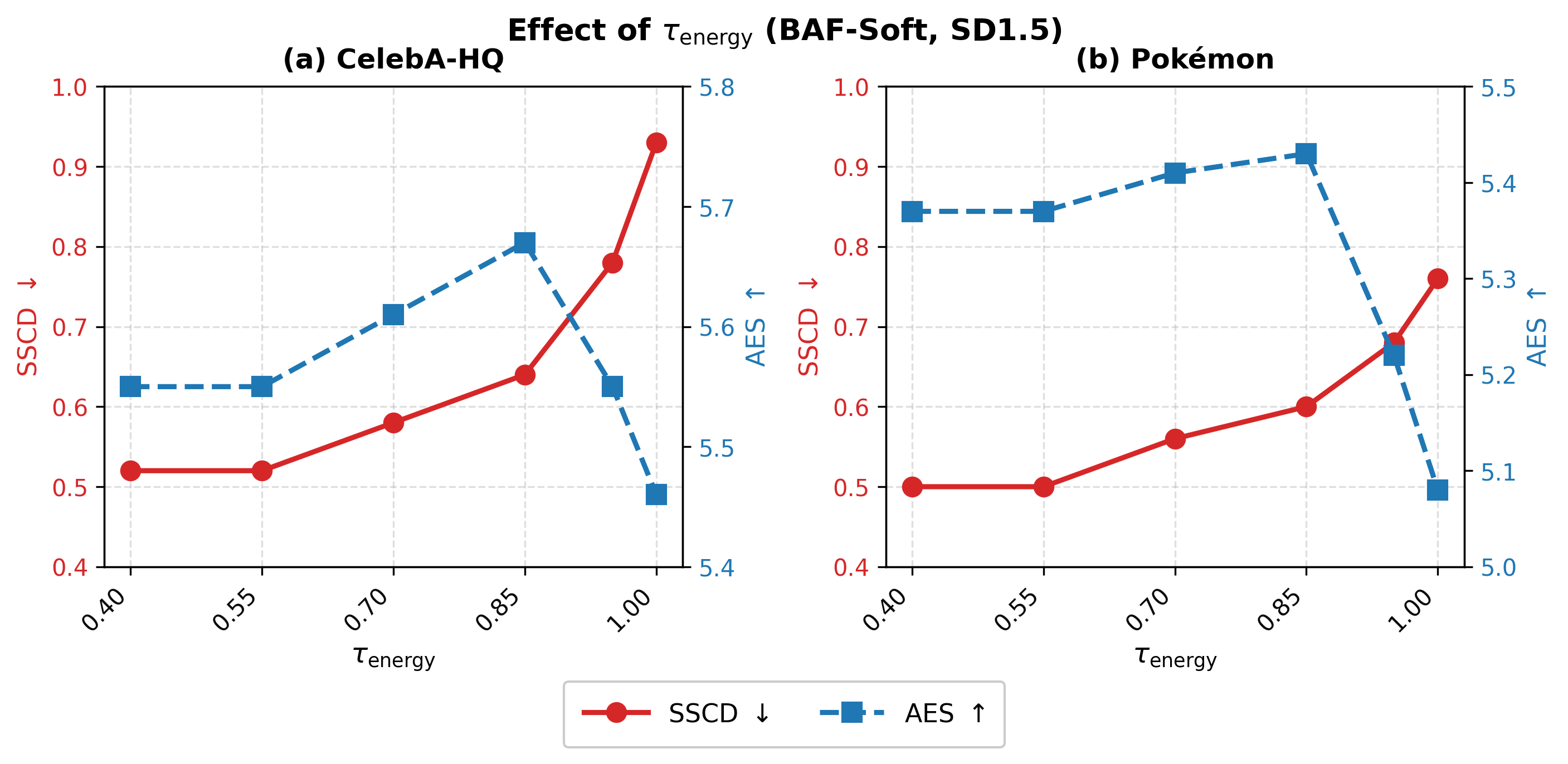}
  \caption{%
    \textbf{Effect of $\tau_\text{energy}$ in BAF-Soft on CelebA-HQ (a) and Pokémon (b), both trained on SD1.5.}
    We measure SSCD (red) and AES (blue). 
  }
  \label{fig:tau_sensitivity}
\end{figure}

Figure~\ref{fig:tau_sensitivity} examines how the energy threshold
$\tau_{\mathrm{energy}}$ affects memorization (SSCD) and generation quality
(AES) on CelebA-HQ and Pokémon. When $\tau_{\mathrm{energy}}$ is small,
memorization is strongly suppressed but AES is slightly reduced,
suggesting that some useful adaptation capacity is also weakened.
As $\tau_{\mathrm{energy}}$ increases, AES improves while SSCD increases
moderately, indicating a better memorization--utility trade-off.
However, when $\tau_{\mathrm{energy}}=1.0$, the filtering effect largely
disappears, leading to a sharp increase in SSCD.

\begin{figure}[t]
  \centering
  \includegraphics[width=\linewidth]{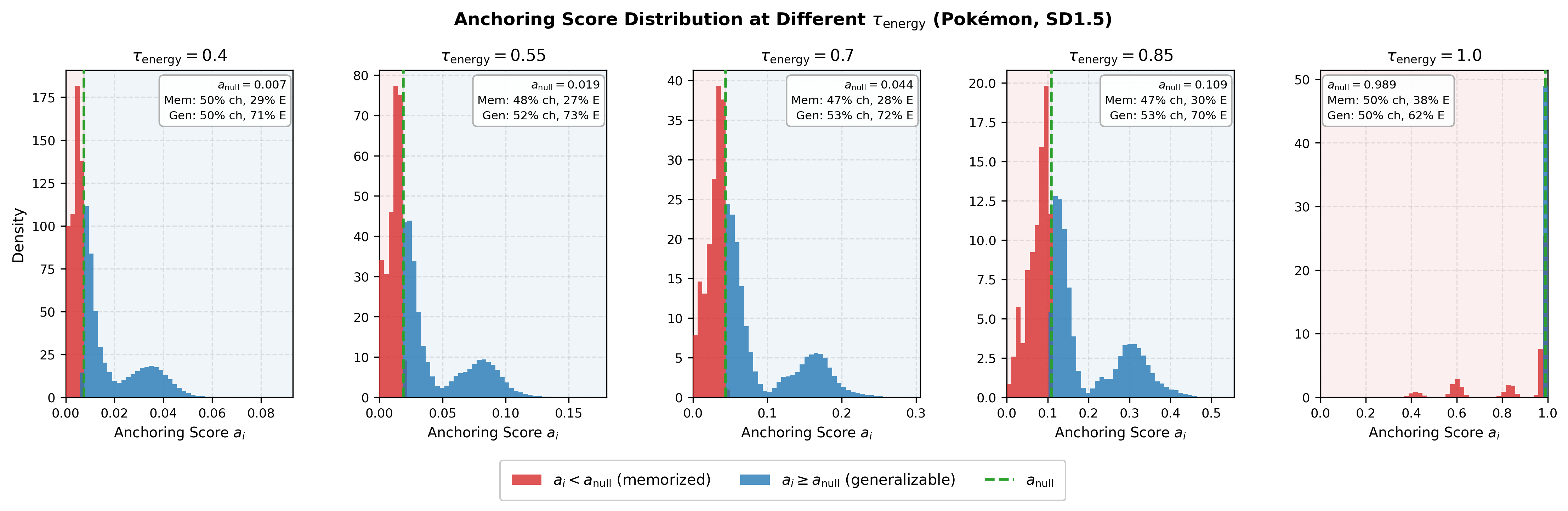}
  \caption{%
    \textbf{Anchoring score distributions of the LoRA at different
    $\tau_{\mathrm{energy}}$ (Pokémon, SD1.5).}
    Each panel shows the distribution of per-channel anchoring scores $a_i$
    for a given $\tau_{\mathrm{energy}}$.
    Red bars indicate channels below the null threshold $a_{\mathrm{null}}$
    (memorized); blue bars indicate channels at or above it (generalizable).
    The green dashed line marks $a_{\mathrm{null}} = (K/m)(K/n)$, where $K$
    is selected by the energy criterion.
    The inset reports the fraction of channels and total energy in each region.
  }
  \label{fig:anchoring_dist}
\end{figure}

To understand the origin of this behavior, we analyze the distribution of
anchoring scores under different thresholds in Figure~\ref{fig:anchoring_dist}. The key factor is that the energy threshold
$\tau_{\mathrm{energy}}$ determines the dimension $K$ of the selected principal subspace of the pretrained model. As $\tau_{\mathrm{energy}}$ increases, more singular directions are included, which enlarges $K$ and therefore raises the null baseline
$a_{\mathrm{null}}=(K/m)(K/n)$.

When $\tau_{\mathrm{energy}}$ is small, the selected principal subspace is very compact. In this regime, the anchoring scores concentrate within a narrow range near zero. Although BAF-Soft weakens many low-alignment directions, the compressed score range limits the ability to reliably distinguish memorized channels from generalizable ones. As $\tau_{\mathrm{energy}}$ increases, the score distribution expands and
becomes increasingly structured. At $\tau_{\mathrm{energy}}=0.85$, a clear
valley emerges between low- and high-alignment regions. Importantly, the
baseline $a_{\mathrm{null}}$ lies close to this valley, allowing BAF-Soft to
suppress low-alignment (memorized) directions while preserving
high-alignment components that contribute to generalizable adaptation. When $\tau_{\mathrm{energy}}=1.0$, the selected subspace spans nearly the
entire spectral space of the pretrained weights. In this case, $K$ approaches
$\min(m,n)$ and $a_{\mathrm{null}}$ becomes close to one. Consequently, most
channels obtain anchoring scores close to the upper end of the range,
causing the soft gating function $g(a_i)=a_i^\alpha$ to approach $1$ for the
majority of channels. The filtering effect therefore largely disappears,
reintroducing memorized directions and leading to the observed increase in
SSCD. These observations suggest that overly small or overly large values of
$\tau_{\mathrm{energy}}$ both weaken the discriminative power of anchoring
scores, while intermediate values provide a more effective separation
between memorized and reusable directions.

\paragraph{Mitigation under Different Memorization Strengths.}
To systematically study memorization behavior, we simulate different memorization strengths by duplicating training images during LoRA
fine-tuning. Specifically, each canary image is repeated $D$ times in the training set, where $D$ denotes the duplication factor. Larger $D$
therefore induces stronger memorization pressure.

To measure memorization under different regimes, we report two complementary
metrics: \textit{targeted SSCD} and \textit{Top-5\% SSCD}. Targeted SSCD
measures similarity between generated images and a specific duplicated
training instance, capturing instance-level memorization of the canary
sample. However, when the duplication factor $D$ is small, memorization may
not concentrate on a particular instance and it becomes unclear which
training image is memorized. In this regime, targeted SSCD becomes
less informative. Hence, we also report \textit{Top-5\% SSCD}, which measures the
$95^{\text{th}}$ percentile similarity between generated images and the
training dataset. This metric is commonly used to detect dataset-level
memorization by identifying whether a small fraction of generated samples
are unusually close to training images.

\begin{figure}[t]
\centering
\includegraphics[width=0.7 \linewidth]{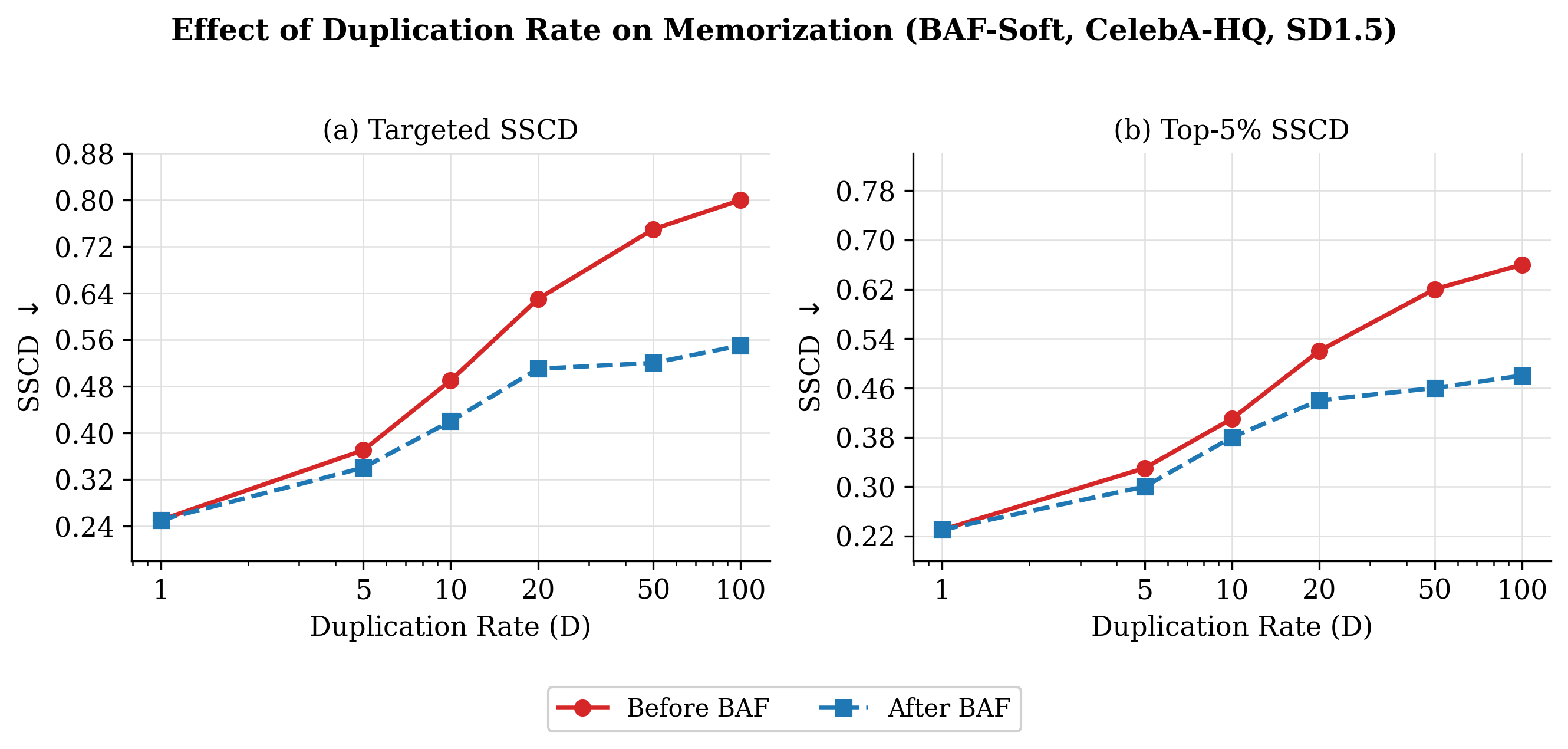}
\caption{
Effect of duplication rate on memorization during LoRA fine-tuning.
To simulate different memorization strengths, each canary image is duplicated $D$ times during training.
The left panel reports \textit{targeted SSCD} and the right panel reports \textit{Top-5\% SSCD}
Red curves denote the model before applying BAF, while blue curves denote the results after applying BAF.
}
\label{fig:duplication_analysis}
\end{figure}

Figure~\ref{fig:duplication_analysis} shows that after applying BAF (blue curves), memorization is consistently reduced
across all duplication levels. Importantly, even when duplication is high
(e.g., $D=50$ or $100$), the resulting SSCD values remain around $0.5$,
which prior work commonly considers to indicate a low likelihood of
memorization. These results demonstrate that BAF effectively suppresses
memorization signals across a wide range of memorization strengths.

\subsection{Real-World Simulation on Civitai LoRAs}
\label{sec:realworld}

The previous experiments use controlled duplication to simulate memorization, following prior protocols~\cite{wen2024detecting}. To assess BAF in deployment-realistic conditions, we further evaluate it on \emph{real copyrighted LoRAs} downloaded directly from Civitai, without access to their training data.
    
    \begin{figure*}[t]
    \centering
    \begin{subfigure}[t]{0.12\linewidth}
        \centering
        \includegraphics[width=\linewidth]{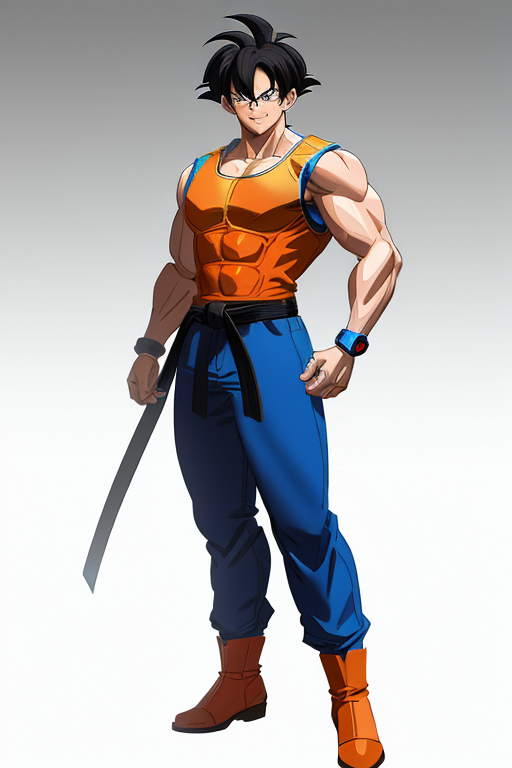}
        \caption{w/o LoRA}
        \label{fig:base}
    \end{subfigure}
    \begin{subfigure}[t]{0.12\linewidth}
        \centering
        \includegraphics[width=\linewidth]{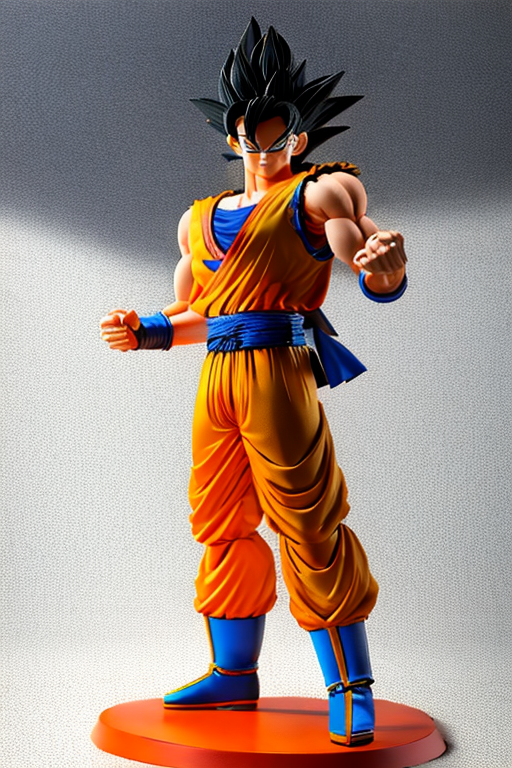}
        \caption{LoRA}
        \label{fig:lora}
    \end{subfigure}
    \begin{subfigure}[t]{0.12\linewidth}
        \centering
        \includegraphics[width=\linewidth]{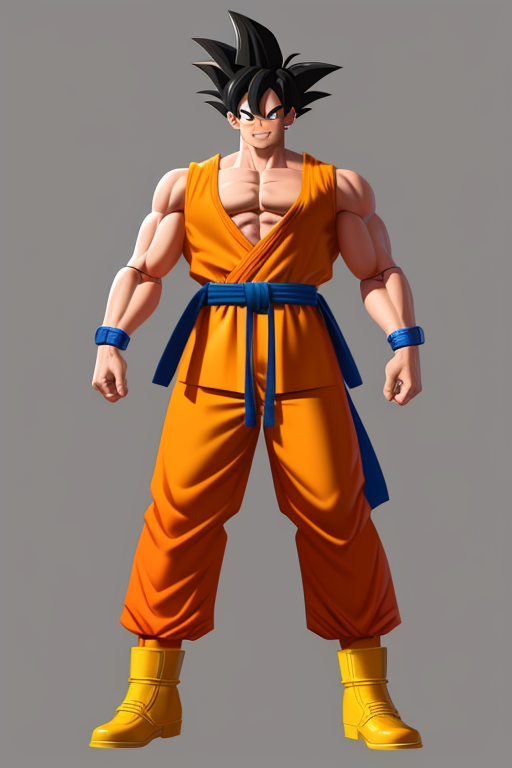}
        \caption{AMG}
    \end{subfigure}
    \begin{subfigure}[t]{0.12\linewidth}
        \centering
        \includegraphics[width=\linewidth]{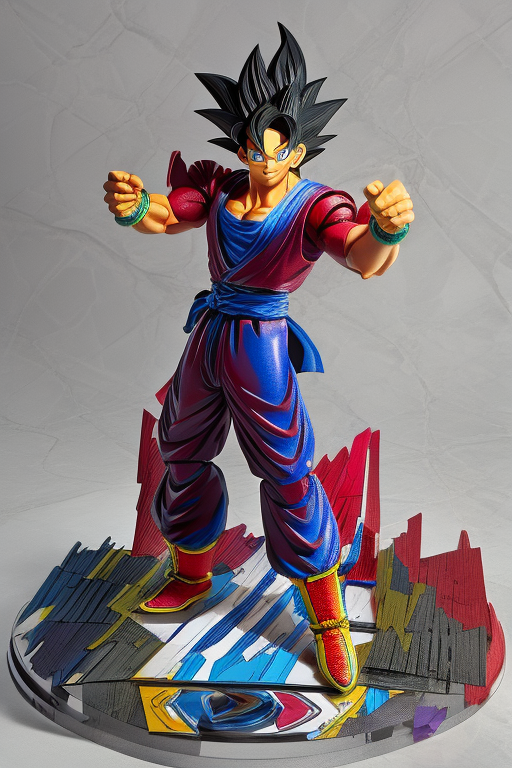}
        \caption{RTA}
    \end{subfigure}
    \begin{subfigure}[t]{0.12\linewidth}
        \centering
        \includegraphics[width=\linewidth]{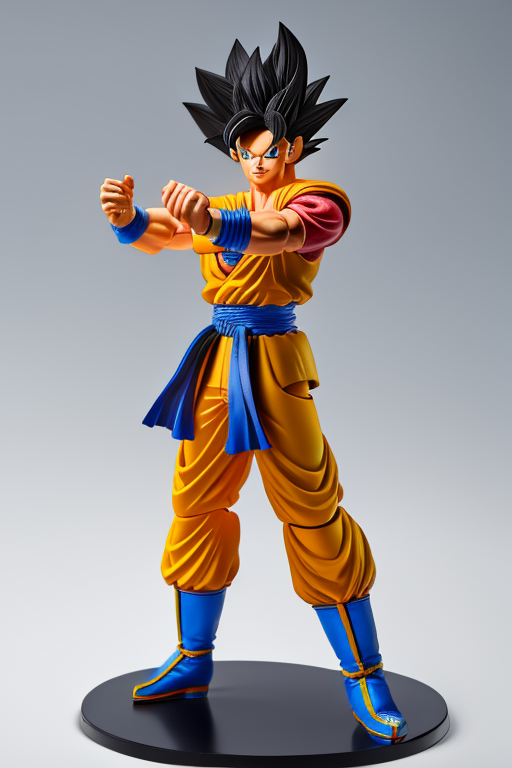}
        \caption{CFG}
    \end{subfigure}
    \begin{subfigure}[t]{0.12\linewidth}
        \centering
        \includegraphics[width=\linewidth]{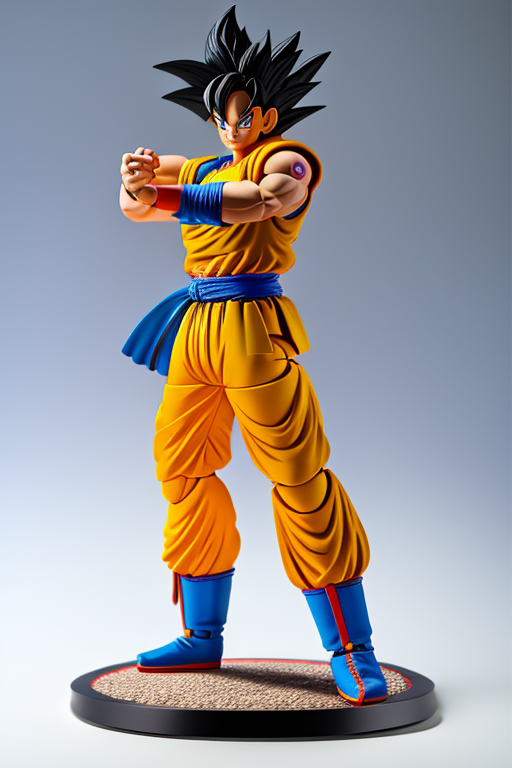}
        \caption{Prompt Optim}
    \end{subfigure}
    \begin{subfigure}[t]{0.12\linewidth}
        \centering
        \includegraphics[width=\linewidth]{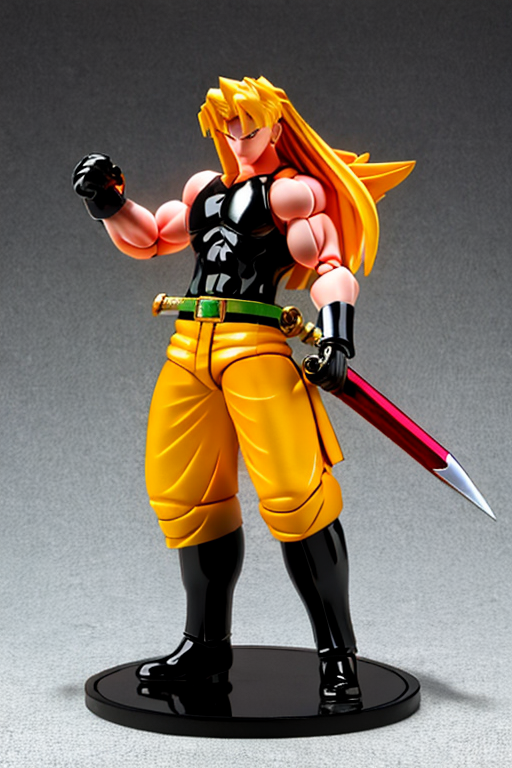}
        \caption{Ours}
    \end{subfigure}
    
    \end{figure*}
    \label{fig:realworld}
    
\paragraph{Setup.} We select 20 publicly distributed Flux LoRAs covering three copyright-sensitive categories: (i) anime characters from popular IPs (e.g., \emph{Frieren}, \emph{Goku}), (ii) celebrity-likeness models, and (iii) IP-specific style LoRAs. For each LoRA, we generate images with identical prompts and seeds before and after applying BAF, and compare against four representative inference-time baselines: AMG, RTA, CFG-Reverse, and Prompt Optim.

\paragraph{Qualitative results.} Figure~\ref{fig:realworld} shows representative results. We compare BAF against existing methods on a Goku Toy-like LoRA. Existing methods either continue to reproduce the copyrighted character near-verbatim or distort the LoRA's toy-like style and visual quality; only BAF yields a visually distinct character while preserving generation quality. Here, we only show one example due to space limit, more results can be found in appendix. We additionally conducted a blind preference study with 10 independent raters unfamiliar with the technical details. For each of the 20 concepts, raters were shown unlabeled outputs from BAF and the four baselines (matched prompt and seed) and asked which method best suppressed reproduction of the copyrighted character. \textbf{BAF was rated as the strongest memorization suppressor in all 20 concepts}, indicating that its advantage observed in the controlled setting carries over to real-world LoRAs.

\section{Conclusion}
In this work, we propose \textbf{BAF (Base-Anchored Filtering)}, a
data-free framework for mitigating memorization directly in the parameter
space of LoRA updates. Experiments on diffusion models demonstrate that BAF consistently reduces
memorization while preserving generation quality across multiple datasets.
Further analysis reveals that anchoring scores exhibit a clear structural
separation between memorized and reusable channels, providing empirical
support for the proposed hypothesis.

Overall, our results suggest that parameter-space structure alone can
provide strong signals for distinguishing memorization from useful
adaptation. We hope this work motivates further research on
\emph{data-free safety mechanisms} for open model ecosystems, where direct
access to training data is often unavailable.

\bibliographystyle{unsrt}  
\bibliography{references}

\end{document}